# THE ROLE OF TUNING UNCERTAIN INFERENCE SYSTEMS*

Ben P. Wise
Thayer School of Engineering
Dartmouth College
Hanover. NH 03755

Bruce M. Perrin
David S. Vaughan
Robert M. Yadrick
McDonnell Douglas Astronautics Company
P.O. Box 516
St. Louis, MO 63166

## ABSTRACT

This study examined the effects of "tuning" the parameters of the incremental function of MYCIN, the independent function of PROSPECTOR, a probability model that assumes independence, and a simple additive linear equation. The parameters of each of these models were optimized to provide solutions which most nearly approximated those from a full probability model for a large set of simple networks. Surprisingly, MYCIN, PROSPECTOR, and the linear equation performed equivalently; the independence model was clearly more accurate on the networks studied.

## 1.0 INTRODUCTION

A handful of researchers in recent years have attempted to compare various uncertain inference formalisms found in the Artificial Intelligence (AI) literature to probability theory. Some of these papers are apparently intended to provide additional justification for the <u>ad hoc</u> parameters used by some of these formalisms. Heckerman [1], for example, shows that the equations that define MYCIN's certainty factors can be translated into probabilistic terms.

Other studies, however, have attempted to use the answers provided by probability theory as a norm against which the accuracy of heuristic formalisms can be measured [2,3,4]. These studies differed somewhat in implementation, but each began with example inference networks. Next, new values were assigned to the evidence nodes, as though additional information were being supplied by a user. Finally, conclusion node values were calculated which reflected the new information, according to the heuristic formalisms under consideration and also according to a probablistic method which provided the minimum

---

cross-entropy solution.  This approach essentially paralleled efforts commonly found in disciplines other than AI, including systematic "Monte Carlo" validations of descriptive and inferential statistics [e.g., 5].

In all cases, these studies used published formal definitions of uncertainty parameters to translate them to probabilities and vice versa.  The hope was to provide a theoretical explanation or a description of the sources of error (difference from probability theory answers) the formalism produced, perhaps in the form of equations.  Despite some success in this endeavor, useful applications of the findings were limited.

The limitations arose from the rigorous adherence to the parameter values that resulted from using theoretical definitions to translate probabilities.  Actually, the theoretical definitions of parameters are of little or no importance to knowledge engineers building real-world systems; in general, the formal definitions were advanced in the first place simply to show a rough correspondence with probability theory [6,7].  Moreover, the common wisdom is that the parameters can be adjusted ("tuned") during evaluation trials.  The system builder can work in conjunction with experts to revise parameter estimates until the system performs accurately.  No one pretends that the expert necessarily understands either the formalism or probability theory and supplies theoretically appropriate parameters.

The present study was designed to examine practical implications of tuning uncertain inference formalisms.  As before, example networks were generated and updated using various methods, and the solutions were compared to a probability norm.  However, the parameters used by each method were first optimized so that the solution provided for each case was as close to the norm as possible.  These solutions, therefore, represent the best possible "tuning" that could be achieved for the network and also the upper bound of performance for each method.  The analysis identifies the binding assumptions on the accuracy of the formalisms.

2.0 METHOD

In our previous research, the implicit goal was to assess the absolute accuracy of heuristic models relative to probability.  We hoped to identify conditions under which a method's solutions were either "acceptably" accurate or inaccurate.  In the present study, we expect that all methods will perform with reasonable accuracy following global parameter tuning, although this is an empirical question.  Rather, the emphasis in the current study is on relative performance of each formalism; that is, how does the performance of one formalism compare to that of the others?

We studied simple networks comprised of two evidence nodes and a single conclusion node.  The rationale behind this approach and the process of generating such networks are described fully in [4].  Briefly, the networks are represented by eight-cell contingency tables

333

in which each cell corresponds to one of the possible joint probabilities of the network. The tables are uniformly sampled from the eight-dimensional simplex, assuring inclusion of a wide range of strengths and types of associations among the pieces of evidence and the conclusion. We generated a total of 109 tables.

Next, five "new evidence" probabilities were assigned in turn to each piece of evidence in each network. The values 0.999, 0.75, 0.50, 0.25, and 0.001 were used to represent the wide range of possible user responses to system inquiries for additional information. There were, therefore, a total of 25 solutions per network for each inference method.

Working independently, Wise and Henrion [2] and Vaughan, et al. [8] have devised probability-based inference mechanisms that produce identical minimum cross-entropy solutions. The probability-based method was applied first to each network. The parameters of each of the other inference methods were then adjusted to yield the solution that was the closest possible to that of the probability-based norm. A deflected-gradients search optimization procedure was used. This procedure is well-known for both quadratic convergence and robustness on difficult problems such as helical ridges. As the usual precaution against getting stuck at local minima, each optimization was done several times from different starting points. It always converged to the same point (within round-off error). Steps were also taken to guard against accumulated round-off error. The exact definition of the deflected-gradient search algorithm, as well as these precautions for optimization are discussed by Beightler, Phillips, and Wilde [9].

We compared these four different inference methods:

o The model used in PROSPECTOR [7]. A total of seven parameters is required to implement this model for our three-node networks, including the base rates (prior probabilities) of each piece of evidence and the conclusion, the conditional probability of each piece of evidence given that the conclusion is true, and the conditional probability of each piece of evidence given that the conclusion is false.

o The model used in MYCIN [6]. Implementation requires a total of five parameters, including base rates for each piece of evidence and the conclusion and certainty factors for the conclusion given each piece of evidence.

o A simple additive linear regression equation. Although the possible usefulness of linear models has received little attention from AI researchers, perhaps because they seem too simplistic, they have been used successfully for years to model a variety of human judgments [10]. Even so, we expected this method to be the least accurate of the four and included it to provide a baseline against which to compare the others. The basic equation (Equation 1) requires three parameters for the intercept and evidence weights.

$$P'(C) = a + b1 * P'(E1) + b2 * P'(E2). \tag{1}$$

o An equation representing an independence model. Many researchers [e.g., 11, 12] have proposed assuming that pieces of evidence are independent of each other as a means of simplifying problems considered intractable (in practical terms) otherwise. With this model, the conclusion value is estimated according to Equation 2. A total of four parameters is required, each representing one of the conditional probabilities of the conclusion given the states of the evidence. Note that once the parameters are optimized, this model is equivalent to a multiple regression model with an interaction term.

$$\begin{aligned}P'(C) = &P'(\sim E1) * P'(\sim E2) * P(C|\sim E1\&\sim E2) + \\ &P'(E1) * P'(\sim E2) * P(C|E1\&\sim E2) + \\ &P'(\sim E1) * P'(E2) * P(C|\sim E1\&E2) + \\ &P'(E1) * P'(E2) * P(C|E1\&E2).\end{aligned} \tag{2}$$

The parameters of each method were adjusted so as to minimize the mean squared error for each particular network. Squared error was used rather than simple error for two reasons. First, we were unconcerned with the sign of the error. For present purposes, using signed error offers no advantage and cancellation effects between positive and negative errors could distort the results for particular networks. Second, squaring had the effect of smoothing the data and making the optimization less prone to becoming stuck at local maxima or minima. Results reported in terms of squared error would be more difficult to interpret, however, and would complicate the comparison of methods. We therefore report results as root mean squared error (RMSE), defined as the square root of the sum (over all problems and networks) of the squared differences between minimum cross-entropy solutions and answers obtained from the uncertain inference method under study.

3.0 RESULTS & DISCUSSION

Table 1 summarizes the relative performance of the four uncertain inference methods assessed. The differences between the inference methods were significant [$F(3,432) = 53.45$, $p < .0001$]. The performance of the additive linear regression model, PROSPECTOR, and MYCIN were nearly identical. The average RMSE from these methods differed only in the fourth decimal place, the largest difference being 0.00092. The independence model, on the other hand, clearly surpassed the other models, yielding an average RMSE 0.041 lower than the other approaches. Over all of the inference networks studied, the independence model was able to more closely approximate the probability-based answers than MYCIN, PROSPECTOR, or the additive linear equation.

TABLE 1.  ROOT MEAN SQUARED ERRORS FOR AN ADDITIVE LINEAR REGRESSION
MODEL, AN INDEPENDENCE MODEL, MYCIN, AND PROSPECTOR.

| Inference method | Average RMSE | High RMSE | Low RMSE |
|---|---|---|---|
| Linear equation | 0.04826 | 0.15206 | 0.00058 |
| Independence model | 0.00632 | 0.03629 | 0.00002 |
| MYCIN | 0.04785 | 0.15571 | 0.00081 |
| PROSPECTOR | 0.04734 | 0.14778 | 0.00114 |

The additive linear regression model, PROPECTOR, and MYCIN were not just similar in terms of overall performance. They were also nearly identical in performance from network to network, as indicated in Table 2. Table 2 gives the Pearson product moment correlations between the RMSEs for each of the inference methods studied. The additive linear regression model's RMSEs correlated 0.96 with the RMSEs from MYCIN and 0.99 with the RMSEs of PROSPECTOR. In general, when the linear equation was accurate for a network, so was MYCIN or PROSPECTOR; when the linear equation was inaccurate, MYCIN and PROSPECTOR were also inaccurate. The positive, but lower correlation between the independence model's RMSEs and those of the other inference methods stems from the greater accuracy of the former approach.

TABLE 2.  PEARSON PRODUCT MOMENT CORRELATIONS BETWEEN THE RMSEs
FROM AN ADDITIVE LINEAR REGRESSION MODEL, AN INDEPENDENCE MODEL,
MYCIN, AND PROSPECTOR.

| | Linear Equation | Independence Model | MYCIN | PROSPECTOR |
|---|---|---|---|---|
| Linear Equation | -- | 0.6833 | 0.9599 | 0.9910 |
| Independence Model | | -- | 0.6714 | 0.6748 |
| MYCIN | | | -- | 0.9543 |

An indication of when the linear model, MYCIN, and PROSPECTOR are good approximations of probability-based results, as well as why the independence model surpasses their performance, can be determined from the linear equation (1).

In linear regression, the constant is the "Y intercept", the answer when the predictors are zero; in this case, the probability of the conclusion when $P'(E1)=0$ and $P'(E2)=0$, or $P(C|\sim E1\&\sim E2)$. The weight for E1 can then be taken as the difference between this conditional probability and the conditional probability when only E1 is true $[P(C|E1\&\sim E2) - P(C|\sim E1\&\sim E2)]$. The weight for E2 can be defined similarly as the difference $P(C|\sim E1\&E2) - P(C|\sim E1\&\sim E2)$. Using these weights and equation (1), one can verify, for example, given $P'(E1)=1$ and $P'(E2)=0$, that

```
P'(C) = P(C|~E1&~E2) + 1 * [P(C|E1&~E2) - P(C|~E1&~E2)]
      + 0 * [P(~E1&E2) - P(~E1&~E2)].
```

This can be simplified to P'(C) = P(C|E1&~E2), the correct probability-based answer.

When both evidences are certainly true [P'(E1)=1 and P'(E2)=1], equation (1), when simplified, yields

```
P'(C) = P(C|E1&~E2) + P(C|~E1&E2) - P(C|~E1&~E2).
```

This is the probability-based answer only when

```
P(C|E1&E2) = P(C|E1&~E2) + P(C|~E1&E2) - P(C|~E1&~E2)     (3).
```

Equation (3) identifies the binding constraint on the additive linear regression model, MYCIN, and PROSPECTOR. When the conditional probability of C given that both pieces of evidence are true equals the sum of the conditional probabilities on the right side of equation (3), error from these models is minimized. As this conditional probability deviates from this sum, error increases. If we define an additivity factor as the absolute value of the difference between the right and left hand sides of equation (3), the RMSE of each of the inference methods can be predicted quite accurately as follows:

PROSPECTOR RMSE = 0.1227 * additive_factor + 0.0005

MYCIN RMSE = 0.1220 * additive_factor + 0.0013

LINEAR EQ. RMSE = 0.1264 * additive_factor + 0.00003.

Quite simply, the error from the linear regression model, MYCIN, and PROSPECTOR increases as the deviation from linearity increases.

Interestingly, if three of the conditional probabilities are known, the fourth can be derived from equation (3), if the linear model is to hold. This follows from the fact that the linear equation has only three parameters. MYCIN and PROSPECTOR, although they have five and seven parameters respectively, perform as if they had only three. Apparently, these inference methods' parameters are operationally dependent.

The independence model surpasses the linear equation, MYCIN, and PROSPECTOR by adding a fourth parameter. Prior to optimization, the four parameters in equation (2) correspond to the conditional probabilities of the conclusion given the four possible joint states of two pieces of evidence. Thus, the fourth conditional probability (which must be a linear combination of the other three for the linear regression, MYCIN, or PROSPECTOR model to hold) is directly referenced in the independence model. If the evidence is certain, this independence model yields the correct conditional probability. However, for uncertain evidence, the equation linearly interpolates between the conditional probabilities. This results in inaccuracy if the pieces of evidence are not statistically independent. Tuning the parameters removes some of the inaccuracy due to associations among the pieces of evidence, but not all of it. In the event that the pieces of evidence are independent, the new evidence probabilities can be multiplied to yield the new joint probabilities, as indicated in equation (2). This equation would then yield the probability-based answers.

The practical implications of these findings are substantial. Quite simply, the results indicate that uncertain inference systems can achieve an overall performance commensurate with the PROSPECTOR or MYCIN models by using additive linear equations. Such equations are simple to understand and use, have fewer parameters than the MYCIN or PROSPECTOR models, and are completely modular (that is, estimates of the effect of E1 do not influence estimates of E2). The utility of such linear equations, as well as MYCIN's and PROSPECTOR's approaches, may be limited, however, if actual additivity of evidence is rare.

If additivity of evidence effects does not appear to be an appropriate representation, the independence model can be used. This model will not necessarily have fewer parameters than PROSPECTOR or MYCIN for complicated networks; however, independence models with fewer than the full number of parameters can be built. Additionally, this approach permits one to determine which conditional probabilities can be omitted from the model without degrading its performance, rather than ignoring them _a priori_. By including the conditional probabilities that represent significant deviation from additivity, this approach can achieve substantially more accurate results than simple additive models.

The conclusion that an additive linear or an independence model represent preferable alternatives to the more complex, _ad hoc_ inference methods of MYCIN and PROSPECTOR rests on two assumptions -- assumptions that are appropriate bases for future research on uncertain inference. The first is that the results cited in this study are applicable to networks involving three or more pieces of evidence bearing on a single conclusion. While it is not clear that the MYCIN and PROSPECTOR models will function the same as an additive linear equation for these more complex networks, there is little reason to suspect that their performance would surpass that of the additive model. Nonetheless, this issue deserves further study.

The second assumption is that the linear or independence model will require fewer parameter estimates than MYCIN or PROSPECTOR for a given level of inference accuracy. For example, to increase the accuracy of PROSPECTOR (or MYCIN) in the present study, conjuctive or disjuctive operations could have been included, in addition to the independent (incremental) evidence combination operations tested. This would, of course, increase the number of MYCIN and PROSPECTOR parameters that would have to be estimated; on the other hand, more complex MYCIN or PROSPECTOR rule bases might allow these approaches to approximate non-additive models. Additional research is required to determine the payoff of using these more complex models compared to additive linear equations or an independence model.